\documentclass{article}
\PassOptionsToPackage{numbers, sort&compress}{natbib}

\usepackage[preprint]{neurips_data_2024}

\usepackage[utf8]{inputenc} 
\usepackage[T1]{fontenc}    
\usepackage{hyperref}       
\usepackage{url}            
\usepackage{booktabs}       
\usepackage{amsfonts}       
\usepackage{nicefrac}       
\usepackage{microtype}      
\usepackage{soul}
\usepackage{multirow}
\usepackage{graphicx}       
\usepackage{makecell}
\usepackage[table,dvipsnames]{xcolor}
\usepackage{amsmath}
\usepackage{subcaption}
\usepackage{hyperref}
\usepackage{wrapfig}
\usepackage{marvosym}
\definecolor{myblue}{RGB}{215,238,247}
\definecolor{mygreen}{RGB}{230,241,221}
\definecolor{mygrey}{RGB}{242,242,242}
\definecolor{myorange}{RGB}{235,142,71}
\definecolor{textgreen}{RGB}{135,201,195}
\definecolor{mypurple}{RGB}{222,213,255}
\definecolor{darkblue}{RGB}{66,141,191}
\definecolor{greyblue}{RGB}{208,220,232}

\title{\textsc{DocBench}: A Benchmark for Evaluating LLM-based Document Reading Systems}

\author{{\bf Anni Zou$^{1,2}$\thanks{This work was done during internship at Tencent AI Lab, Seattle.}} ,  Wenhao Yu$^{2}$\textsuperscript{\Letter},  Hongming Zhang$^{2}$,  Kaixin Ma$^{2}$, \\ 
\bf Deng Cai$^{2}$,  Zhuosheng Zhang$^{1}$,  Hai Zhao$^{1}$,  Dong Yu$^{2}$ \\
$^{1}$Shanghai Jiao Tong University \
$^{2}$Tencent AI Lab \
\\
\texttt{anni0103zou@gmail.com}, \\ \texttt{\textsuperscript{\Letter}wenhaowyu@global.tencent.com}
\text{(corresponding author)}
}

\begin{document}
\maketitle

\begin{abstract}

Recently, there has been a growing interest among large language model (LLM) developers in LLM-based document reading systems, which enable users to upload their own documents and pose questions related to the document contents, going beyond simple reading comprehension tasks. Consequently, these systems have been carefully designed to tackle challenges such as file parsing, metadata extraction, multi-modal information understanding and long-context reading. However, no current benchmark exists to evaluate their performance in such scenarios, where a raw file and questions are provided as input, and a corresponding response is expected as output.
In this paper, we introduce \textsc{DocBench}, a new benchmark designed to evaluate LLM-based document reading systems. Our benchmark involves a meticulously crafted process, including the recruitment of human annotators and the generation of synthetic questions. It includes 229 real documents and 1,102 questions, spanning across five different domains and four major types of questions.
We evaluate both proprietary LLM-based systems accessible via web interfaces or APIs, and a parse-then-read pipeline employing open-source LLMs.
Our evaluations reveal noticeable gaps between existing LLM-based document reading systems and human performance, underscoring the challenges of developing proficient systems.
To summarize, \textsc{DocBench} aims to establish a standardized benchmark for evaluating LLM-based document reading systems under diverse real-world scenarios, thereby guiding future advancements in this research area.~\footnote{Data and code will be released at \url{https://github.com/Anni-Zou/DocBench}.}

\end{abstract}

\vspace{-0.05in}
\section{Introduction}

The emergence of large language models (LLMs) has marked a significant milestone in the field of natural language processing, revolutionizing the way we approach a variety of tasks~\cite{zhao2023survey,chang2024survey,wang2024survey,achiam2023gpt,anthropic2024claude,touvron2023llama2,team2023gemini}. Existing LLMs such as GPT-4~\cite{achiam2023gpt}, Llama-3~\cite{touvron2023llama2}, and Claude-3~\cite{anthropic2024claude} have shown exceptional abilities in following human instructions to perform tasks such as answering questions, translating languages and summarizing texts. These tasks are typically characterized by straightforward input-output interactions, where the models generate responses solely based on the provided text. 
However, many real-world applications require more complex interactions involving user-provided documents. For instance, financial analysts might need to query comprehensive financial reports to inform their investment decisions~\cite{liu2021finbert,yang2023fingpt,wu2023bloomberggpt}. Legal professionals often search through extensive legal documents to find relevant case law~\cite{lai2023large,chen2024survey,cui2023chatlaw}. Similarly, scientific researchers frequently sift through academic papers to identify related works and extract key findings~\cite{dasigi2021dataset,birhane2023science}.

\begin{figure}[t]
\centering
\includegraphics[width=1.0\textwidth]
{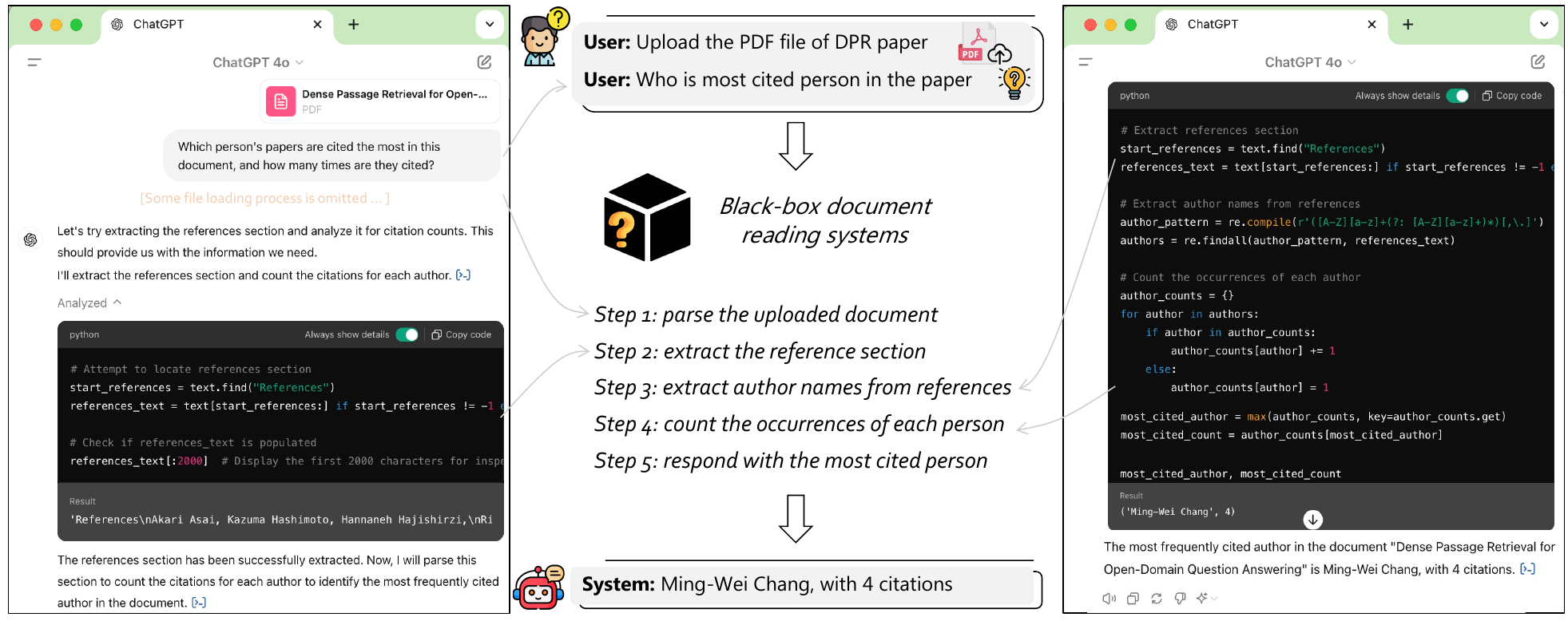}
\caption{An example of OpenAI's GPT-4 based document reading system. Unlike standalone LLMs, recent proprietary LLM-based document reading systems employ a carefully designed approach (e.g., file parsing, code execution) to answer user questions related to document contents.}
\vspace{-0.1in}
\label{fig:example}
\end{figure}

When users pose queries based on their provided documents, the situation becomes more intricate and challenging~\cite{lee2024human}. Unlike standalone LLMs that are primarily trained to process and respond to textual inputs (or images in the case of Vision LLMs), handling user-provided documents necessitates a more sophisticated approach that stretches beyond the capabilities of a single LLM. 
In order to provide accurate responses, an LLM-based document reading system should not only comprehend natural language queries, but also excel in a range of processing skills, including parsing and interpreting user documents and layouts, navigating complex formatting structures, extracting relevant metadata, and managing long textual contexts along with any embedded images. Mastery of these diverse skills is essential for generating precise and contextually relevant responses.



At the same time, recent advancements in proprietary LLM developers such as OpenAI and Anthropic have provoked the release of several LLM-based document reading systems. Figure \ref{fig:example} illustrates an example of OpenAI’s GPT-4-based document reading system. 
Despite widespread claims of effectiveness and efficiency in various online public blogs\footnote{Blog: Claude can now use tools \url{https://www.anthropic.com/news/tool-use-ga}}\footnote{Blog: Talk with documents using LlamaIndex \url{https://codemaker2016.medium.com/talk-with-documents-using-llamaindex-3952c76bd511}}, \textbf{the absence of a standardized benchmark} makes it difficult to objectively evaluate and compare the document reading performance across these systems, thereby leaving a critical gap in fairly assessing these capabilities in a fine-grained manner.


To fill this gap, our paper introduces \textsc{DocBench}, a novel benchmark specifically designed to evaluate LLM-based document reading systems. \textsc{DocBench} is developed to mirror real-world scenarios where each input consists of a document paired with one or multiple associated questions, and each question is annotated with a golden answer. Our benchmark undergoes a meticulous development process, incorporating human annotation and synthetic question generation. To the end, \textsc{DocBench} features 229 real-world documents and 1,102 questions spanning 5 diverse domains: \emph{Academia, Finance, Government, Laws, and News}. Besides, the benchmark involves 4 question categories, including \textit{text-only, multi-modal (i.e., tables and figures), meta-data, and unanswerable}, ensuring comprehensive coverage of various document reading capabilities.

Based upon \textsc{DocBench}, we evaluate several proprietary LLM-based systems that are accessible via web interfaces or APIs. However, these proprietary systems are close-sourced, thus leading to the limited disclosure of their detailed operational strategies. As a result, we additionally assess a straightforward parse-then-read pipeline employing a series of open-source LLMs.
Our evaluations reveal noticeable gaps between existing LLM-based document reading systems and human performance, underscoring the challenges of developing proficient systems.

In summary, \textsc{DocBench} serves as the first standardized benchmark to evaluate LLM-based document reading systems within real-world scenarios, where the systems take a document file paired with one or multiple related questions as input and generate textual responses as output. Moreover, our benchmark is carefully designed to encompass 5 diverse domains and 4 distinct question types, ensuring a nuanced and thorough assessment. By facilitating fair comparisons across different systems, \textsc{DocBench} highlights current limitations and paves the way for future advancements.

\vspace{-0.1in}
\section{The \textsc{DocBench}}

\textsc{DocBench} is a benchmark that takes raw \texttt{PDF} files and accompanying questions as inputs, with the objective of generating corresponding textual answers.
In this section, we will introduce the pipeline used to construct the dataset, present detailed statistics, and explain the evaluation method.


\vspace{-0.05in}
\subsection{Dataset Construction}
\vspace{-0.05in}

Our dataset construction pipeline consists of three phases. First, we crawl documents across various domains from publicly accessible online resources ($\S$\ref{sec:doc_collection}). Second, we generate corresponding QA pairs with the help of GPT-4 and a team of human annotators ($\S$\ref{sec:qa_generation}). Finally, we conduct auto filtering followed by a manual review to validate the quality of the generated instances ($\S$\ref{sec:qa_check}). 

\begin{figure}[t]
\centering
\includegraphics[width=1.0\textwidth]{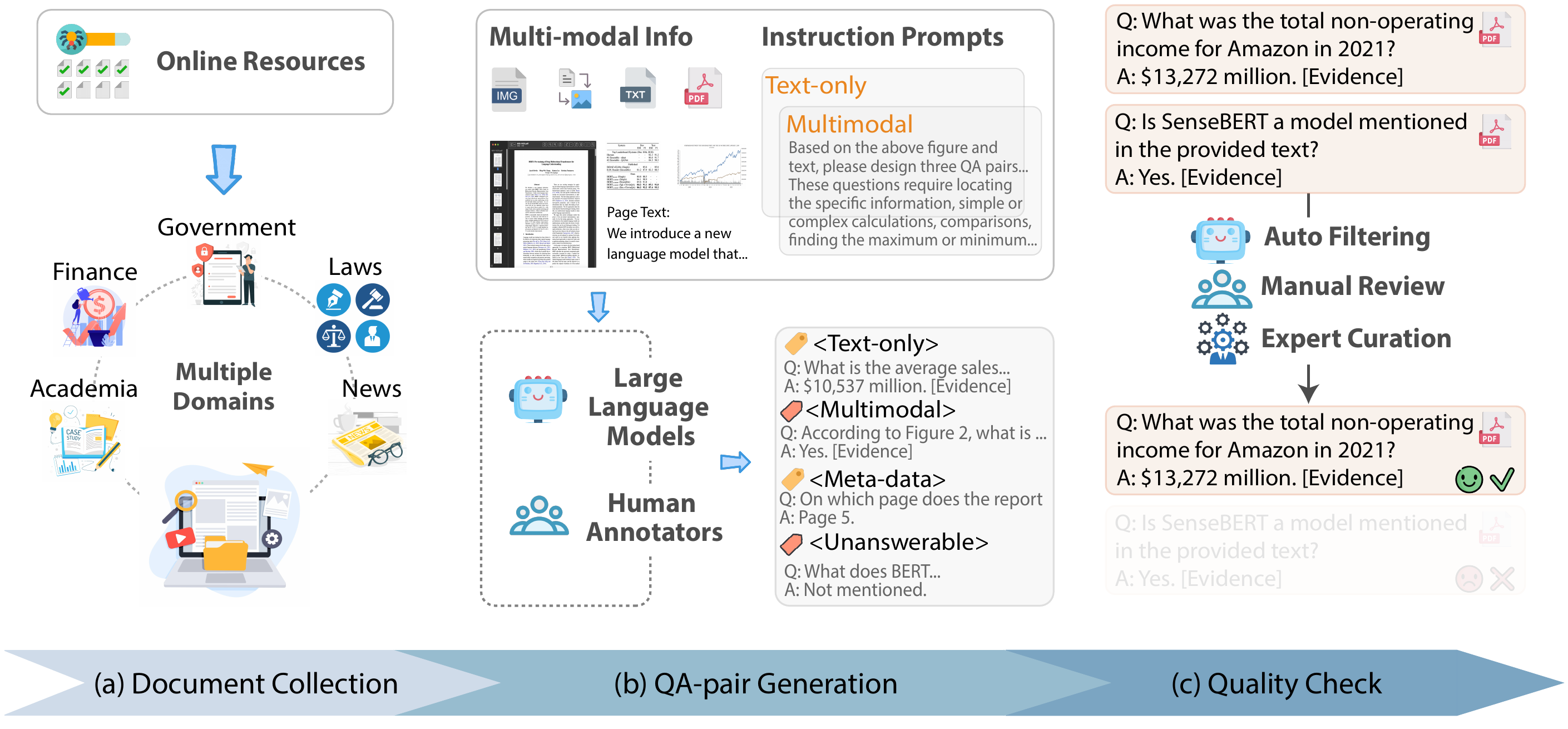}
\caption{Construction pipeline of \textsc{DocBench}. (a) Document Collection: gathering \texttt{PDF} files from five different domains; (b) QA-pair Generation: creating diverse and comprehensive QA pairs through a combination of LLMs and human effort; (c) Quality Check: ensuring data quality through a multi-step process that includes auto filtering, manual review, and expert curation.}
\vspace{-0.1in}
\label{fig:construction}
\end{figure}

\vspace{-0.05in}
\subsubsection{Document Collection}
\label{sec:doc_collection}
\vspace{-0.05in}

To establish a practical and constructive benchmark for document reading, we concentrate on scenarios where it is crucial to read documents. We standardize the documents to \texttt{PDF} format due to its high compatibility and stability. We identify five domains where documents are frequently utilized: \emph{Academia}, \emph{Finance}, \emph{Government}, \emph{Laws}, \emph{News}. For \underline{Academia}, papers are downloaded from arXiv within the range of top-$k$ citations in the field of natural language processing on Google Scholar.~\footnote{\url{https://scholar.google.com/}; \url{https://arxiv.org/}.} For \underline{\emph{Finance}}, we crawl the annual reports of companies with top-$k$ global market capitalization up to \texttt{2024-02-23} from AnnualReports.~\footnote{\url{https://companiesmarketcap.com}; \url{http://www.annualreports.com}.} For \underline{\emph{Government}}, we manually download official governmental reports in 2023 from the U.S. Department of State and GovInfo.~\footnote{\url{https://www.state.gov/department-reports/}; \url{https://www.govinfo.gov/}.} For \underline{\emph{Laws}}, files are gathered from an official online collection of publications from the Library of Congress, within the years ranging from 2020 to 2024.~\footnote{\url{https://www.loc.gov/collections/publications-of-the-law-library-of-congress}.} For \underline{\emph{News}}, we collect front-page scanned documents of the New York Times, covering dates from \texttt{2022-02-22} to \texttt{2024-02-22}.~\footnote{\url{https://static01.nyt.com/images/}.} 
We set $k=100$ in the initial crawling process for academic and financial documents. After skipping the unobtainable or damaged documents, we eventually obtained 229 \texttt{PDF} files, with 49 for academia, 40 for finance, 44 for government, 46 for laws, and 50 for news. Detailed statistics are shown in Table \ref{tab:data-statistics}.






\vspace{-0.05in}
\subsubsection{QA-pair Generation}
\label{sec:qa_generation}
\vspace{-0.05in}

The generation procedure revolves around two aspects: diversity and comprehensiveness.
On one hand, as the document itself inherently abounds with multi-dimensional and multi-modal information including texts, tables, figures, and meta-data, we leverage the \texttt{fitz} library~\footnote{\url{https://pypi.org/project/fitz/}} to parse out the distinct modalities within the \texttt{PDF} files. Afterward, we deliver plain texts to GPT-4 (\texttt{gpt-4-0125-preview}) for generating \emph{text-only} QA pairs and resort to GPT-4V (\texttt{gpt-4-1106-vision-preview}) for yielding multi-modal ones based on tables, figures, and their related textual descriptions.
On the other hand, we further request a set of human annotators to manually elaborate 350 QA pairs based on the given document files. Their primary task is to focus on types that are rarely covered in the previous generation stage but are frequent in daily usage, such as meta-data and unanswerable instances.
Details and additional analysis of instruction prompts are attached in Appendix \ref{app:instruction_prompt}.

\vspace{-0.05in}
\subsubsection{Quality Check}
\label{sec:qa_check}
\vspace{-0.05in}

We begin by instructing GPT-4 to automatically filter out questions that are excessively lengthy, unnatural, or impractical. We then conduct a manual review following the automatic filtering to ensure both the quality of questions and the accuracy of answers. To further align our data with real-world user scenarios, we engage 7 practitioners from distinct domains to review and refine the data within their areas of expertise. In this way, our data quality is validated from multiple perspectives.

\vspace{-0.05in}
\subsection{Dataset Statistics}
\vspace{-0.05in}

\begin{table*}[t]
\footnotesize
\setlength{\tabcolsep}{15pt}
\renewcommand\arraystretch{1.0}
\centering
\caption{Overview statistics of \textsc{DocBench}. All documents are in \texttt{PDF} format. We extract text content and calculate the corresponding \emph{\#Tokens} of documents.}
\vspace{-0.1in}
\scalebox{0.9}{\begin{tabular}{lrr|rrrr}
\toprule
    \multirow{2}{*}{\textbf{Category}} & \multicolumn{2}{c}{\textbf{Questions.}} & \multicolumn{4}{c}{\textbf{Documents.}} \\ 
\cmidrule(l){2-3} \cmidrule(l){4-7} 
        &\textbf{\#Num} & \textbf{\#Tokens} & \textbf{\#Num} & \textbf{\#Pages} & \textbf{\#Size(KB)} & \textbf{\#Tokens}  \\ 
\midrule
Aca.    &303    &16.8   &49     &11        &847         &11,123     \\
Fin.    &288    &16.8   &40     &192       &6,594       &149,409    \\
Gov.    &148    &14.1   &44     &69        &2,183       &36,105     \\
Laws    &191    &15.4   &46     &58        &969         &32,339     \\
News    &172    &13.5   &50     &1         &3,095       &2,909      \\
\midrule
Total/Avg. &1,102 &15.7 &229    &66        &2,738       &46,377     \\ 
\bottomrule
\end{tabular}}
\vspace{-0.05in}
\label{tab:data-statistics}
\end{table*}

\begin{figure}[t]
\centering
\includegraphics[width=1.0\textwidth]{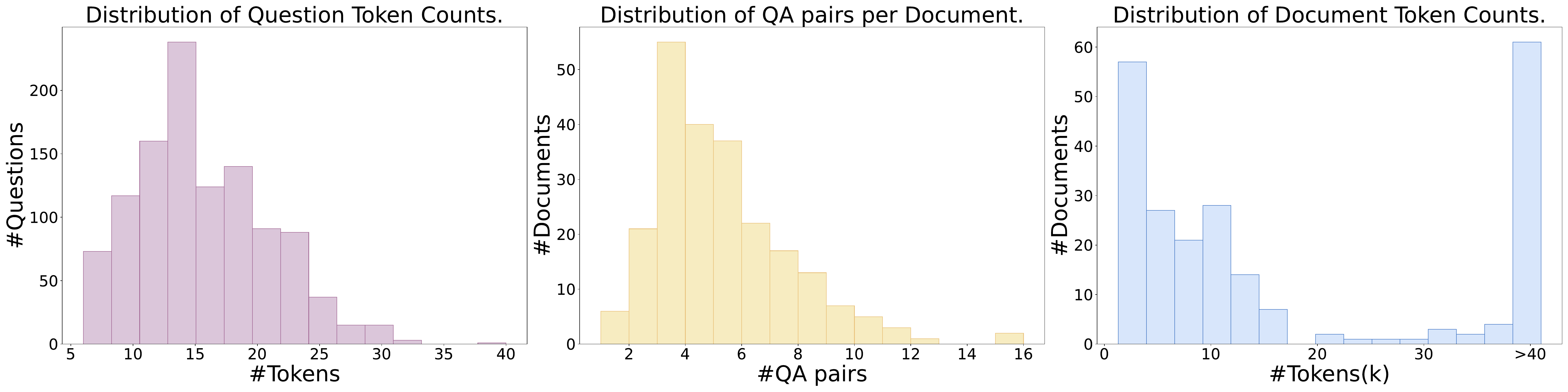}
\vspace{-0.23in}
\caption{Overview of Questions and Documents: distribution of question token counts (left); distribution of QA pairs per document (middle); distribution of document token counts (right).}
\vspace{-0.1in}
\label{fig:dist_hist}
\end{figure}

\textsc{DocBench} comprises a total of 229 \texttt{PDF} documents sourced from publicly accessible online repositories along with 1,102 questions, spanning across 5 domains: Academia, Finance, Government, Law, and News. 
As shown in Table \ref{tab:data-statistics}, we conduct comprehensive statistical analysis across various angles, encompassing the number of questions, documents, and average token counts within each. Given the unique nature of our task input, which involves processing \texttt{PDF} files, we additionally include information such as page count and file size. Moreover, Figure \ref{fig:dist_hist} presents distributions depicting the counts of question tokens, document tokens~\footnote{We utilize the tokenizer of \texttt{gpt-4-turbo} for token measurement.}, and QA pairs per document. Notably, we constrain the number of QA pairs per document to a maximum of $20$, with its range spanning from $1$ to $16$, aiming to better emulate real-world usage scenarios. As for the token counts of questions and documents, the minimum and maximum values are $(6 || 40)$ and $(1,300 || 598,302)$ respectively.

\begin{table*}[t]
\centering
\setlength{\tabcolsep}{2.5mm}
\caption{Examples of instances from \textsc{DocBench}, with multiple labels indicating our data diversity.}
{\scalebox{0.83}{
\begin{tabular}{p{4cm}p{3.5cm}p{2.3cm}p{5.1cm}}
\toprule
\textbf{Question} & \textbf{Answer} & \textbf{Labels} & \textbf{Document} \\
\midrule
\textbf{\textcolor{myorange}{Why}} does the model not perform as well in German compared to Spanish and Dutch? & Due to its \textbf{\textcolor{darkblue}{complex morphology and compound words}}... & \texttt{<Aca.><Why> \newline <Text-only> <Textual>} & When and Why are Pre-trained Word Embeddings Useful
for Machine Translation \href{https://arxiv.org/pdf/1804.06323}{[clickable file link]}\\ 
\midrule
By \textbf{\textcolor{myorange}{how much}} did the number of Erica users increase from 2018 to 2019? & The number increased by \textbf{\textcolor{darkblue}{5.5 million}}... &  \texttt{<Fin.><How> \newline <Multimodal> <Numerical>}  & Bank of America Annual Report 2020
\href{https://www.annualreports.com/HostedData/AnnualReportArchive/b/NYSE_BAC_2020.pdf}{[clickable file link]}\\
\midrule
\textbf{\textcolor{myorange}{What}} is the primary focus of Bureau Objective 3.4? & The report \textbf{\textcolor{darkblue}{does not contain}} such objective. &  \texttt{<Gov.> <Wh-> \newline <Unanswerable> <Others>} &  Governmental report from \emph{Secretary’s Office of Global Women’s Issues} 2022
\href{https://www.state.gov/wp-content/uploads/2022/02/S_GWI_FBS_FINAL_Public-Version-1.pdf}{[clickable file link]}\\
\midrule
\textbf{\textcolor{myorange}{How many}} times does the report mention "scientific ethics"? & The report mentions "scientific ethics" \textbf{\textcolor{darkblue}{11}} times. &  \texttt{<Laws><How> \newline <Meta-data> <Numerical>} &  Report on \emph{Regulation of Stem Cell Research} from Library of Congress 2023 \href{https://tile.loc.gov/storage-services/service/ll/llglrd/2023555925/2023555925.pdf}{[clickable file link]}\\ 

\midrule
\textbf{\textcolor{myorange}{Is}} the article about Hurricane Ian's impact in Florida written by multiple authors? & \textbf{\textcolor{darkblue}{Yes}}, the article is about Hurrican Ian's impace in Florida... &
\texttt{<News><Y/N> \newline <Meta-data> <Boolean>} & New York Times front page on \texttt{2022-09-30} \href{https://static01.nyt.com/images/2022/09/30/nytfrontpage/scan.pdf}{[clickable file link]}\\
\bottomrule
\end{tabular}}}
\vspace{-0.1in}
\end{table*}

\begin{figure}[t]
\centering
\includegraphics[width=0.97\textwidth]{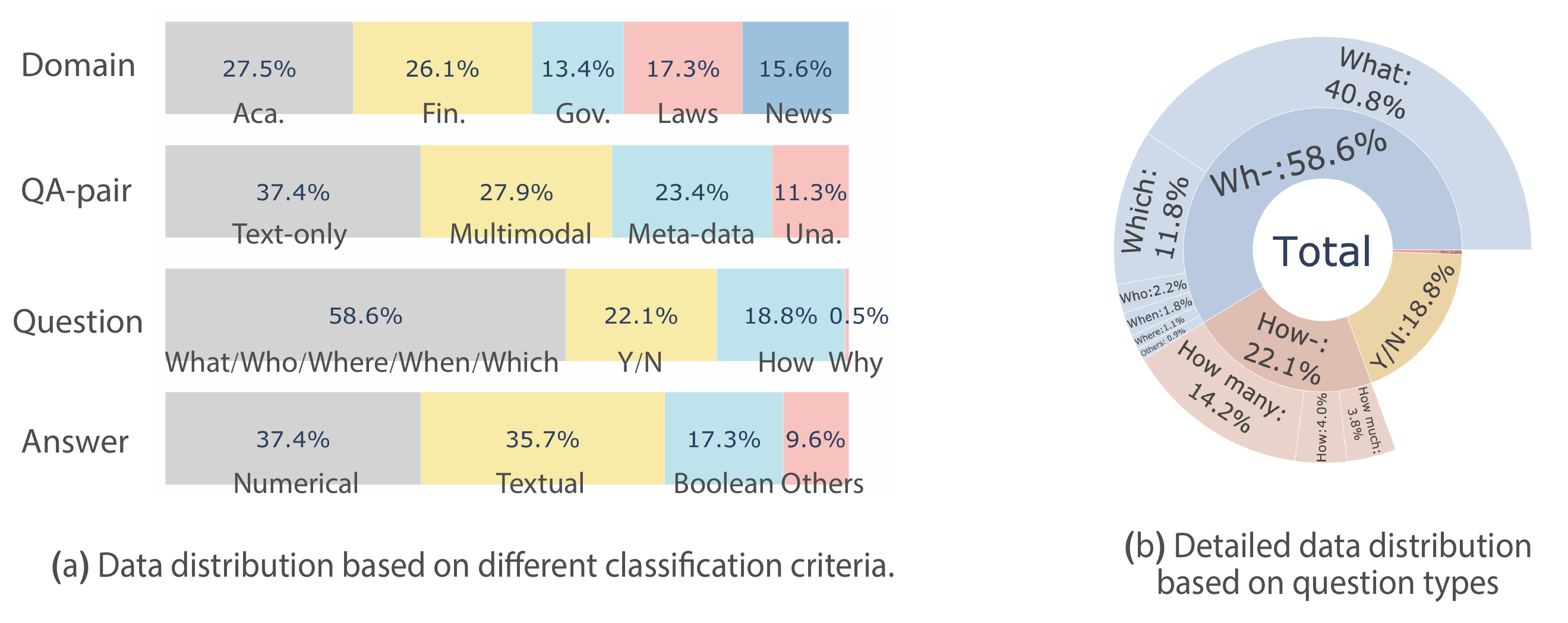}
\vspace{-0.1in}
\caption{Data distribution of \textsc{DocBench}: (a) proportion(\%) of various data groups based on four distinct classification criteria; (b) detailed data analysis based on question types.}
\label{fig:distribution}
\vspace{-0.1in}
\end{figure}

\vspace{-0.05in}
\subsection{Dataset Analysis}\label{sec:data_anal}
\vspace{-0.05in}

Figure \ref{fig:distribution} illustrates the data distribution in \textsc{DocBench} based on different classification criteria.

\vspace{0.03in}
\noindent\textbf{QA-pair Type} The types of QA pairs can be mainly divided into four groups: \emph{text-only} (37.4\%), \emph{multimodal} (27.9\%), \emph{meta-data} (23.4\%), and \emph{unanswerable} (11.3\%).
The \emph{text-only} and \emph{multimodal} types collectively account for over half (65.3\%), centering on the abilities to comprehend long contexts and interpret information from different modalities. Besides, we incorporate approximately one-third (34.7\%) of questions to more closely fit the actual scenarios as well as assess the robustness of the document reading systems, including 23.4\% inquiring about metadata (e.g., page numbers, word counts) and 11.3\% that cannot be answered based on the given document.

\vspace{0.03in}
\noindent\textbf{Question Type}
The types of questions can be primarily separated into four categories according to the inquiry focus: \emph{what / who / where / when / which} (58.6\%), \emph{Y/N} (22.1\%), \emph{how} (18.8\%), and \emph{why} (0.5\%). 
These categories respectively demand specific information or details, straightforward \emph{yes} or \emph{no} responses, methods or degrees, and the underlying reasons behind actions or phenomena.
Figure \ref{fig:distribution}(b) delineates a detailed data distribution based on question types. The interrogative \emph{what} holds a dominant proportion at 40.8\%, which is reasonable as users commonly seek precise information when confronted with a document.

\vspace{0.03in}
\noindent\textbf{Answer Type}
The types of answers can be partitioned into four classes: \emph{numerical} (37.4\%), \emph{textual} (35.7\%), \emph{boolean} (17.3\%), and \emph{others} (9.6\%). 
Within the \emph{numerical} class, 69\% originate from the domains of \emph{academia} and \emph{finance}, as these documents naturally require extensive use of numbers to convey information, such as performance metrics in academic papers and figures in financial reports.

\subsection{Evaluation Setup}

\vspace{-0.03in}
\noindent\textbf{Evaluation Process} \quad
Our dataset diversity poses two major evaluation challenges: 
(i) The evaluation methods vary depending on the answer type. For example, for boolean or numerical answers, a fair evaluator only needs to verify the correctness of a binary \emph{yes/no} response or a specific number using simple techniques like string matching or number extraction. In contrast, textual responses require more nuanced standards such as natural language generation (NLG) metrics. Thus, accurately determining the appropriate evaluation method becomes complex when the answer type is unknown.
(ii) Different LLMs and systems exhibit substantial variations in the organization and style of their outputs, potentially leading to biases in traditional evaluation approaches.
Therefore, we capitalize on the prowess of LLMs that have proven to be decent evaluators and can be easily adapted to the assessment of various answer types~\citep{fu2023gptscore,liu-etal-2023-g,wang-etal-2023-chatgpt}.
Inspired by \citet{liu-etal-2023-g}, we clearly define the evaluation criteria for various types within the instruction prompt and then instruct GPT-4 to assign a score of 0 (incorrect) or 1 (correct).
After evaluating 200 examples by both human evaluators and GPT-4, we found that the GPT-4 automatic evaluator shows a 98\% agreement with human annotators, significantly exceeding the traditional string matching approach. Details of this experiment is shown in Table \ref{tab:metric}, and details of evaluation instruction prompts are attached in Appendix \ref{app:instruction_prompt}.

\vspace{-0.03in}
\noindent\textbf{Metrics} \quad
As mentioned above, we instruct GPT-4 to assign a score of 0 (incorrect) or 1 (correct), thus using Accuracy (abbreviated as $\mathrm{Acc.}$) to measure system performance. 
We report accuracy across all instances, as well as for each domain and QA-pair type in Table \ref{tab:main_results}. 


\begin{table*}[t]
\footnotesize
\setlength{\tabcolsep}{5pt}
\renewcommand\arraystretch{0.95}
\centering
\caption{The GPT-4 automatic evaluator shows a 98\% agreement with human annotators. We randomly sample 40 questions and answers from five systems, asking human annotators to assess their accuracy. We then employ string matching (StrMatch), GPT-3.5, and GPT-4 as automatic evaluators. Finally, we measure the agreement between the human and these automatic evaluators. }
\vspace{-0.05in}
\resizebox{\columnwidth}{!}{%
\begin{tabular}{l|cccc|ccc}
\toprule
\multirow{2}{*}{\textbf{Sources}} &\multicolumn{4}{c|}{\textbf{\# Correct / Wrong by different evaluators}} &\multicolumn{3}{c}{\textbf{Agreement (human and automatic evaluators)}}  \\
\cmidrule(lr){2-5} \cmidrule(lr){6-8} 
&\textbf{Human} &\textbf{GPT-4} &\textbf{GPT-3.5} &\textbf{StrMatch} &\hspace{4ex} \textbf{GPT-4} &\hspace{4ex} \textbf{GPT-3.5} &\textbf{StrMatch} \\ 
\midrule
KimiChat & 24 / 16 & 23 / 17 & 33 / \ 7 & 0 / 40 & \hspace{4ex} 97.5\% & \hspace{4ex} 75.0\% &  40.0\% \\
Qwen-2.5 & 17 / 23 & 18 / 22 & 31 / \ 9 & 0 / 40 & \hspace{4ex} 97.5\% & \hspace{4ex} 57.5\% &  57.5\% \\
Gemma (7B) & 19 / 21 & 18 / 22 & 18 / 22 & 0 / 40 & \hspace{4ex} 97.5\% & \hspace{4ex} 75.0\% &  52.5\% \\
Mixtral (7B) & 14 / 26 & 14 / 26 & 26 / 14 & 0 / 40 & \hspace{4ex} 100.0\% & \hspace{4ex} 65.0\% &  65.0\% \\
Llama-3 (70B) & 16 / 24 & 15 / 25 & 28 / 12 & 0 / 40 & \hspace{4ex} 97.5\% & \hspace{4ex} 62.5\% &  60.0\% \\ 
\midrule
Total & 90 / 110 & 88 / 112 & 136 / 64 & 0 / 200 & \hspace{4ex} 98.0\% & \hspace{4ex} 67.0\% & 55.0\% \\
\bottomrule
\end{tabular}
}
\vspace{-0.15in}
\label{tab:metric}
\end{table*}

\begin{table*}[htbp]
\centering
\caption{ Results on \textsc{DocBench} across various types and domains. \emph{Ver./Size} stands for the model version or size; \emph{File} denotes the maximum uploaded file size; \emph{Cxt.} refers to model's context length.}
\renewcommand\tabcolsep{3.1pt}
\renewcommand\arraystretch{1.45}

\label{tab:main_results}
\resizebox{\linewidth}{!}{%
\begin{tabular}{lccc|ccccccccc|c}
\toprule
\multirow{2}{*}{\textbf{Methods}} & \multirow{2}{*}{\textbf{Form}} & \multirow{2}{*}{\makecell[c]{\textbf{Ver.}\\ \textbf{/Size}}} & \multirow{2}{*}{\makecell[c]{\textbf{File}\\ \textbf{/Cxt.}}} & \multicolumn{5}{c}{\textbf{Domain}} & \multicolumn{4}{c|}{\textbf{Type}} & \multirow{2}{*}{\textbf{Overall $\mathrm{Acc.}$}} \\ 
\cmidrule(lr){5-9} \cmidrule(lr){10-13} 
& & &  &\textbf{Aca.} & \textbf{Fin.} & \textbf{Gov.} & \textbf{Laws} & \textbf{News} &\textbf{Text.} & \textbf{Multi.} & \textbf{Meta.} & \textbf{Una.} & \\
\midrule
Human &-  &- &-  &83.0 &82.2 &77.8 &75.0 &86.4 &81.4 &83.3 &77.5 &82.2 &81.2 \\ \\[-1.5em]
\rowcolor{greyblue} \multicolumn{14}{c}{\emph{~~~~~~LLM-based systems}}  \\ \\[-1.1em]
GPT-4     &\emph{API}   &\texttt{0409} &\texttt{100M} &\underline{65.7} &\textbf{65.3} &\underline{75.7} &69.6 &79.6 &\textbf{87.9} &\textbf{74.7} &\underline{50.8} &37.1 &\underline{69.8}  \\ 
GPT-4o    &\emph{API}   &\texttt{0513} &\texttt{100M} &56.4 &56.3 &73.0 &65.5 &75.0 &85.0 &62.7 &50.4 &17.7 &63.1  \\ 
GLM-4     &\emph{Web} &-  &\texttt{20M} &55.8 &35.4 &61.5 &62.8 &82.0 &73.1 &50.3 &48.8 &33.1 &56.5 \\ 
KimiChat      &\emph{Web} &- &\texttt{100M}  &62.4 &\underline{61.8} &\textbf{77.0} &\underline{78.5} &\textbf{87.2} &\underline{87.6} &\underline{65.3} &50.4 &\textbf{71.8} &\textbf{70.9} \\ 
Claude-3  &\emph{Web} &\texttt{Opus} &\texttt{10M} &\textbf{73.9} &40.6 &70.3 &\textbf{79.1} &\underline{86.6} &80.8 &64.6 &\textbf{54.3} &\underline{58.9} & 67.6  \\ 
Qwen-2.5  &\emph{Web} &- &\texttt{150M}  &42.9 &29.9 &51.4 &55.5 &69.2 &61.7 &31.8 &36.0 &58.1 &46.9  \\ 
ERNIE-3.5 &\emph{Web} &- &\texttt{10M} &56.4 &37.5 &54.7 &58.1 &58.1 &63.6 &47.7 &36.8 &54.0 &51.8  \\ 
\midrule\midrule
\rowcolor{greyblue} \multicolumn{14}{c}{\emph{~~~~~~Parse-then-Read Pipelines}}  \\ \\[-1.1em]
GPT-4   &\emph{API}  &\texttt{0409} &\texttt{128k}  &\textbf{70.0} &\textbf{47.9} &\textbf{68.9} &\textbf{70.7} &\textbf{93.6} &\textbf{79.1} &\textbf{63.3} &\textbf{54.3} &\underline{70.2} &\textbf{67.9}  \\
GPT-3.5   &\emph{API}  &\texttt{0125} &\texttt{16k} &49.8 &24.0 &58.8 &50.3 &83.7 &65.0 &37.0 &42.6 &44.4 &49.6  \\
ChatGLM3   &\emph{Open}  &\texttt{6B} &\texttt{128k} &34.7 &41.7 &58.1 &51.3 &58.1 &70.4 &40.3 &31.0 &12.1 &46.2 \\
Gemma     &\emph{Open}  &\texttt{7B} &\texttt{8k} &34.3 &12.5 &43.2 &34.0 &65.1 &43.0 &17.2 &21.3 &\textbf{77.4} &34.6 \\
Mixtral    &\emph{Open}  &\texttt{7B} &\texttt{32k} &42.6 &29.2 &58.8 &50.3 &82.0 &71.8 &33.8 &38.4 &30.6 &48.7  \\
InternLM2   &\emph{Open}  &\texttt{7B} &\texttt{32k} &38.6 &27.1 &52.0 &46.1 &65.7 &63.3 &28.9 &35.3 &25.8 &42.9  \\
Llama-3     &\emph{Open}  &\texttt{8B} &\texttt{8k} &44.6 &23.6 &61.5 &54.5 &86.6 &68.0 &29.2 &45.0 &49.2 &49.6 \\
Yi-1.5     &\emph{Open}  &\texttt{9B} &\texttt{16k} &40.6 &26.4 &58.1 &52.4 &83.1 &66.0 &33.8 &45.7 &27.4 &47.9  \\
Llama-2     &\emph{Open}  &\texttt{13B} &\texttt{4k} &20.8 &18.4 &29.7 &23.6 &55.2 &43.4 &15.9 &21.7 &12.9 &27.2 \\
Phi-3   &\emph{Open}  &\texttt{14B} &\texttt{128k} &50.2 &\underline{44.4} &65.5 &\underline{64.4} &76.7 &77.4 &45.8 &45.3 &44.4 &\underline{57.4}  \\
InternLM2   &\emph{Open}  &\texttt{20B} &\texttt{32k} &43.2 &28.5 &59.5 &54.5 &80.8 &73.3 &33.4 &43.0 &22.6 &49.4 \\
Yi-1.5     &\emph{Open}  &\texttt{34B} &\texttt{16k} &47.2 &27.1 &59.5 &56.5 &78.5 &68.2 &39.0 &49.2 &19.4 &50.1  \\
Command-R   &\emph{Open}  &\texttt{35B} &\texttt{128k} &49.5 &38.9 &66.2 &\underline{64.4} &80.8 &\underline{78.4} &\underline{50.0} &\underline{49.6} &13.7 &56.4  \\
Mixtral-8x7B     &\emph{Open}  &\texttt{47B} &\texttt{32k} &48.5 &31.9 &60.1 &59.2 &81.4 &76.0 &42.9 &46.9 &12.1 &52.7  \\
Llama-3     &\emph{Open}  &\texttt{70B} &\texttt{8k} &\underline{52.1} &25.3 &\underline{68.2} &59.2 &\underline{90.7} &69.2 &38.6 &49.2 &56.5 &54.5 \\
\bottomrule
\end{tabular}
}
\end{table*}

\vspace{-0.10in}
\section{Experiments and Analysis}
\vspace{-0.05in}

\subsection{Experimental Setup}
\vspace{-0.05in}

We conduct a comprehensive evaluation of 22 LLM-based document reading systems, encompassing both proprietary systems that support document uploads and a series of \emph{parse-then-read} pipelines. 
For \emph{parse-then-read} pipelines, we leverage the \texttt{fitz} package to extract text and image blocks from \texttt{PDF} files. We retain the original texts and line breaks for text chunks while we denote the $i$-th image as \emph{[image i]} for images.
Our selection for the proprietary systems includes GPT-4 and GPT-4o~\citep{achiam2023gpt} from OpenAI, GLM-4~\footnote{\url{https://chatglm.cn/main/doc}} from ZhipuAI, Kimi~\footnote{\url{https://kimi.moonshot.cn}} from Moonshot AI, Claude-3~\footnote{\url{https://claude.ai/chats}} from Anthropic, Qwen-2.5~\footnote{\url{https://tongyi.aliyun.com/qianwen}} from Alibaba Cloud, and ERNIE-3.5~\footnote{\url{https://yiyan.baidu.com}} from Baidu.
In the case of the \emph{parse-then-read} pipelines, we assess 15 prominent LLMs as base models, featuring those from the GPT~\citep{achiam2023gpt,gpt3.5}, Llama~\citep{touvron2023llama2}, Mistral~\citep{jiang2024mixtral}, Yi~\citep{young2024yi}, InternLM~\citep{cai2024internlm2}, Phi-3~\citep{abdin2024phi}, Gemma~\citep{team2024gemma}, ChatGLM3~\citep{du2021glm}, and Command-R~\citep{commandr} families. The selection of base open-sourced LLMs adheres to three guiding principles:
(i) official release with \emph{instruct} or \emph{chat} versions 
 that are supported by vLLM~\citep{kwon2023efficient} framework;
(ii) model sizes ranging from 7B to 70B to accommodate GPU memory constraints;
(iii) availability of the longest context length and the latest version.

\subsection{Results and Discussion}
Table \ref{tab:main_results} showcases the performance of various document reading systems on \textsc{DocBench}. 
Our findings reveal substantial variations in document reading capabilities among these systems, driven by differences in their foundational models, context length limitations, diverse design and implementation approaches, and etc. 
In this section, we will provide further discussions to delve deeper into the pros and cons of existing systems, as well as uncover the core challenges posed by \textsc{DocBench}.

\subsubsection{Interpreting Multi-modal and Metadata Information}
\begin{figure}[t]
\centering
\includegraphics[width=1.0\textwidth]{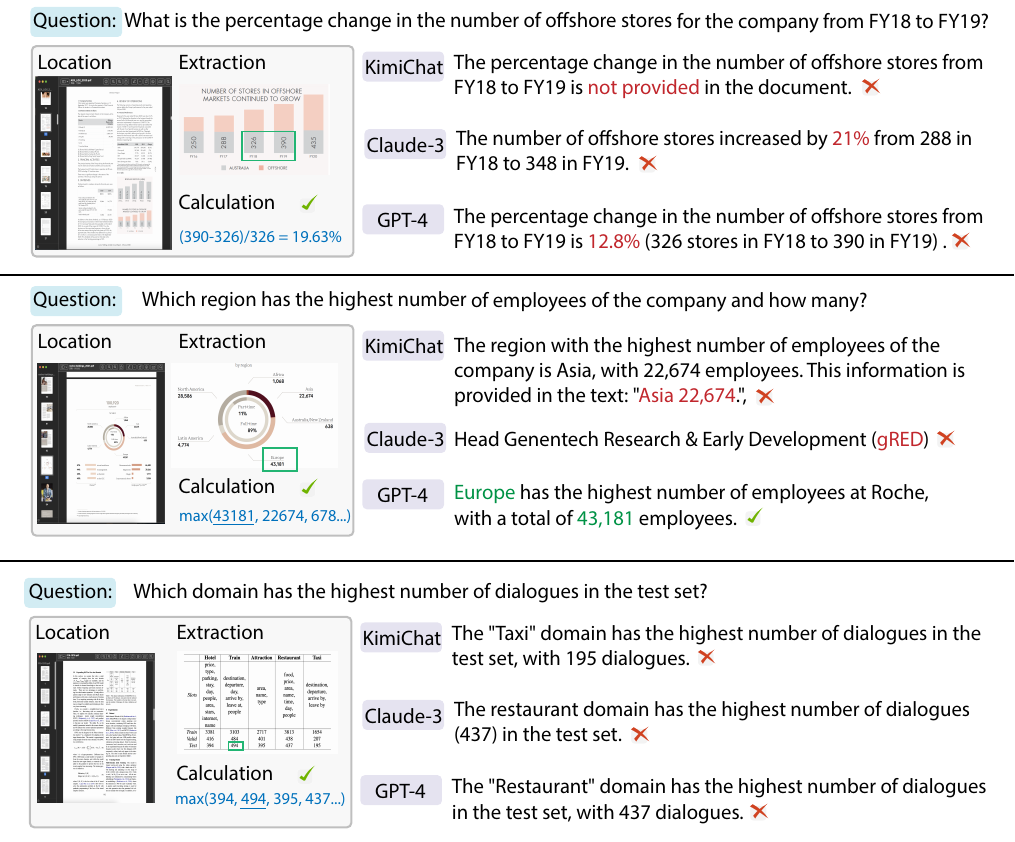}
\caption{To address multi-modal questions in \textsc{DocBench}, it is essential to: (i) identify the relevant figure/table (Location); (ii) extract specific data (Extraction); (iii) perform necessary calculations (Calculation). In the first case study, KimiChat fails to locate the figure, Claude-3 retrieves incorrect data, and GPT-4, despite succeeding in the first two steps, struggles with the calculation.}
\label{fig:case}
\end{figure}

Figure \ref{fig:case} presents a case study illustrating the unique challenge of answering multi-modal questions in \textsc{DocBench}.
We observe that \textbf{leading proprietary LLM-based systems often fail due to errors in one of the steps in the \emph{Location$\rightarrow$Extraction$\rightarrow$Calculation} sequence.}
Take the first case study as an example, in the first step, KimiChat fails to locate the relevant chart on page 17. In the extraction phase, Claude-3 misidentifies the data as \emph{288 \& 348}, instead of the correct \emph{326 \& 390}. Finally, while GPT-4 locates and extracts the correct information, it errs in calculating the percentage change, demonstrating the complexity of these questions.
Interestingly, parse-then-read pipelines can achieve reasonable performance on multi-modal questions (e.g., 63.3\% for GPT-4). This is likely because the parsing process captures certain table information, and documents often include textual descriptions of figures.
Meanwhile, for metadata-related questions, \textbf{current methods generally lack attention to global information}, resulting in relative low performances (below 55\%).

\subsubsection{Handling Lengthy Documents}

\begin{wrapfigure}[14]{r}{0.5\textwidth}
      \centering
      \vspace{-0.1in}
      \includegraphics[width=0.48\textwidth]{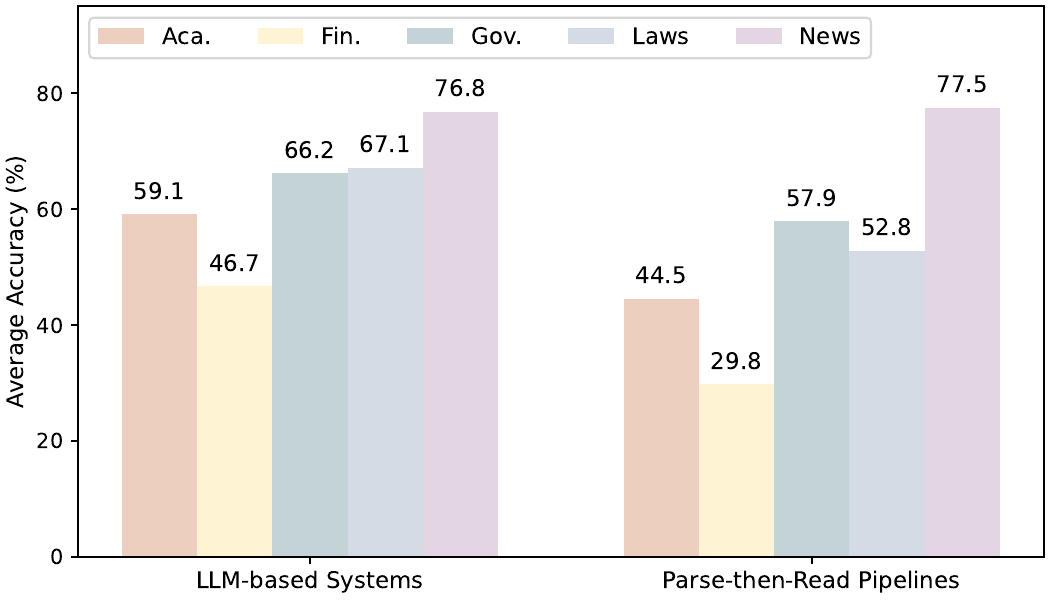}
      \caption{Average accuracy (\%) of two methods under five different domains.}
      \label{fig:domain}
      \vspace{-3.5mm}
\end{wrapfigure}
\textbf{Handling lengthy documents is demanding, especially in real-world scenarios where document size can be virtually unlimited.} Proprietary LLM-based systems struggle with uploading extensive files, while the parse-then-read pipelines with open-sourced LLMs are constrained by their maximum context length, leading to varying degrees of information loss.
As shown in Figure \ref{fig:domain}, both methods perform poorly in the \underline{finance} domain but achieve higher performance in the \underline{news} domain. This discrepancy arises because financial documents are typically longer and contain richer information, whereas news files are limited to single front pages with fewer messages. 
Furthermore, certain strong models with relatively short context lengths may excel with smaller files, but context length becomes a crucial factor when it comes to large files. For instance, the \emph{8k} Llama-3 family performs exceptionally well in the \underline{news} domain, but is outperformed by all the \emph{128k} models in the \underline{finance} domain.
Besides, we discover that KimiChat and Command-R, which are specifically enhanced for long-context and Retrieval-Augmented Generation (RAG) capabilities, achieve decent results on \emph{text-only} questions.
\textbf{Therefore, a key challenge lies in adapting these systems to handle documents of varying lengths while balancing the foundational model's capabilities and context length constraints.}

\subsubsection{Faithfulness to User-provided Documents}

\textbf{Most existing document reading systems falter when faced with unanswerable questions based on the provided document, exhibiting a lack of fidelity.}
Remarkably, Gemma and KimiChat perform better in such scenarios, which represents a crucial capability since users often expect systems to answer questions strictly based on given files. 
Intriguingly, despite the commonly-shared base model on GPT-4, there is a notable performance gap between the system and the parse-then-read pipeline in handling unanswerable questions (i.e., 37.1\% and 70.2 \% for system and pipeline, respectively). We analyze that this may be due to: (i) the proprietary LLM-based system have undergone optimizations on the base model, potentially causing overfitting; (ii) GPT-4 tends to adhere more closely to the in-context learning information. \textbf{Such phenomenon thus underscores a critical challenge for future document reading systems on enhancing fidelity to the given documents.}
 

\section{Related Works}

\subsection{Recent Advances of LLMs and LLM-based Systems}

The latest generation of LLMs, such as GPT-4~\cite{achiam2023gpt}, Llama-3~\cite{touvron2023llama2} and Claude-3~\cite{anthropic2024claude},  have significantly extended the capabilities of language models~\cite{zhao2023survey,chang2024survey,wang2024survey}. 
These models are pre-trained on vast amounts of web-scale data, enabling them to perform a wide range of human-instructed tasks with impressive performance.
Despite their remarkable performance, standalone LLMs may not be sufficient for many real-world applications. For example, LLMs lack access to real-time information and may struggle with tasks that require up-to-date knowledge~\cite{vu2023freshllms}. Moreover, real-world applications often require non-text inputs parsing, code execution, API calling and interaction with external environments ~\cite{lee2024human,cognition2024devin,jimenez2023swebench,zhou2023webarena,xie2024osworld,guo2024dsagent}. The overall task completion usually requires multiple reasoning, execution and reflection steps that cannot be accomplished in a simple input-output manner~\cite{yao2023react,shinn2023reflexion,wang2024executable}. 
To overcome the limitations of standalone LLMs, recent efforts have incorporated additional components and sophisticated system design. These systems, such as Microsoft's Co-Pilot\footnote{\url{https://copilot.microsoft.com}} and OpenAI's GPT-4 all-in-one\footnote{\url{https://chat.openai.com}}, aim to provide more comprehensive and practical solutions for real-world applications. Other pioneering efforts on designing LLM-based systems include web agents~\cite{zheng2024gpt4vision,he2024webvoyager,ma2023laser}, software agents~\cite{yang2024swe,cognition2024devin} and computer agents~\cite{wu2024oscopilot} that can interact with external resources (e.g., websites, search engine, code repositories or computers) and perform multi-step tasks. The success of these systems relies on integrating powerful LLMs with well-designed architectures and components that enable them to handle complex tasks effectively.

\vspace{-0.05in}
\subsection{Document reading: Datasets and Methods}
Document reading is a critical area where LLM-based systems have demonstrated significant advancements. Proprietary developers such as OpenAI\footnote{OpenAI's ChatGPT: \url{https://chat.openai.com}} and Anthropic\footnote{Anthropic's Claude: \url{https://claude.ai/chats}} have introduced advanced systems that can take a user-provided document as input, parse its structure, extract relevant metadata, and handle long texts and images to provide accurate responses. 
While these systems build upon the fundamental capabilities of their underlying LLMs~\cite{zeng2022glm,bai2023qwen,achiam2023gpt,anthropic2024claude}, they differ in their design and implementation, with some systems excelling in long-context reading and others focusing on retrieval-augmented methods to improve document reading ability. 
Despite claims of effectiveness and efficiency in online public blogs, the absence of a standardized benchmark makes it difficult to objectively evaluate and compare the document reading performance across these systems.
Existing benchmarks relevant to document reading are unable to adequately reflect the real performance of these systems. Datasets focusing on document understanding such as Doc2Dial~\cite{feng-etal-2020-doc2dial}, ConditionalQA~\cite{sun-etal-2022-conditionalqa} and those specifically focusing on long-context reading like NarrativeQA~\cite{kovcisky2018narrativeqa} and QuALITY~\cite{pang2022quality}, primarily use text as input only, ignoring the complex nature of document structure and multi-modal information. On the other hand, multi-modal document reading datasets like DocVQA~\cite{mathew2021docvqa}, ChartQA~\cite{masry2022chartqa}, OCR-VQA~\cite{mishra2019ocr}, and InfoVQA~\cite{mathew2022infographicvqa} include multi-modal inputs and preserve the original document structure and layout. However these datasets often capture only parts of document (e.g. tables or figures) and ignored substantial amount of textual content. Different from previous works, DocBench requires systems to process the full documents as intact files and covers different types of questions targeting various abilities, which can more accurately evaluate the capabilities of LLM-based document reading systems in real-world scenarios. 


\section{Conclusion}
In this paper, we introduce \textsc{DocBench}, a novel benchmark created to assess LLM-based document reading systems in a comprehensive and fine-grained manner. \textsc{DocBench} consists of 229 documents and 1,102 questions, spanning 5 domains and 4 question types, developed with the help of human annotators and synthetic questions. We evaluate both proprietary LLM systems, accessible via web interfaces or APIs, and a parse-then-read approach using open-source LLMs. Our findings reveal significant disparities in document reading capabilities among these systems, highlighting current limitations, presenting potential challenges, and thus driving forward progress in this research field.

\bibliographystyle{ACM-Reference-Format}
\bibliography{references}

\appendix

\section{Instruction Prompts}\label{app:instruction_prompt}
\subsection{Response Evaluation}
Detailed instruction prompts for response evaluation are shown in Table \ref{tab:prompt_eval}.
\begingroup
\begin{table*}[ht]
    \centering
    \caption{Instruction Prompts in Response Evaluation.\label{tab:prompt_eval}}
    \vspace{2.8mm}
    \begin{tabular}{p{0.96\linewidth}}
        \toprule
        \sethlcolor{myblue}\hl{\textbf{\texttt{System Content:}}} \\
         \quad You are a helpful evaluator. \\
         
        \\

        \\

        \sethlcolor{mygreen}\hl{\textbf{\texttt{Prompt:}}} \\
        \textbf{Task Overview:} \\
        \quad You are tasked with evaluating user answers based on a given question, reference answer, and additional reference text. Your goal is to assess the correctness of the user answer using a specific metric.\\ 
        \\
        \textbf{Evaluation Criteria:} \\
        \quad 1. Yes/No Questions: Verify if the user's answer aligns with the reference answer in terms of a "yes" or "no" response.\\ 
        \quad 2. Short Answers/Directives: Ensure key details such as numbers, specific nouns/verbs, and dates match those in the reference answer.\\
        \quad 3. Abstractive/Long Answers: The user's answer can differ in wording but must convey the same meaning and contain the same key information as the reference answer to be considered correct.\\
        \\
        \textbf{Evaluation Process:} \\
        \quad 1. Identify the type of question presented.\\ 
        \quad 2. Apply the relevant criteria from the Evaluation Criteria.\\
        \quad 3. Compare the user's answer against the reference answer accordingly.\\
        \quad 4. Consult the reference text for clarification when needed.\\
        \quad 5. Score the answer with a binary label 0 or 1, where 0 denotes wrong and 1 denotes correct.\\
        \quad NOTE that if the user answer is 0 or an empty string, it should get a 0 score.\\
        \\
        \textbf{Question:} \texttt{\{\{question\}\}}\\
        \textbf{User Answer:} \texttt{\{\{sys\_ans\}\}} \\
        \textbf{Reference Answer:} \texttt{\{\{ref\_ans\}\}} \\
        \textbf{Reference Text:} \texttt{\{\{ref\_text\}\}} \\
        \\
        \textbf{Evaluation Form (score ONLY):}\\
        \quad - Correctness: \\      
\bottomrule
\end{tabular}
\end{table*}
\endgroup
\subsection{QA-pair Generation}
Details of instruction prompts for generating QA pairs are attached in Table \ref{tab:prompt_qa}.
We discover that simply passing diagrams to GPT-4V leads to subpar question quality. This issue likely stems from the fact that figures or tables without accompanying text descriptions typically lack sufficient information, thus causing the generated QA pairs to deviate from their intended meanings. In addition, we observe that adding difficulty settings for QA generation (e.g., \emph{Easy}, \emph{Medium}, \emph{Hard}) in the instruction prompt can result in higher quality. We analyze that this may be due to the model being able to favor higher generation quality in potential comparisons.
\begingroup
\begin{table*}[ht]
    \centering
    \caption{Instruction Prompts in QA-pair Generation.\label{tab:prompt_qa}}
    \vspace{2.8mm}
    \begin{tabular}{p{0.96\linewidth}}
        \toprule
        \sethlcolor{myblue}\hl{\textbf{\texttt{System Content:}}} \\
         \quad You are a helpful assistant that can generate question-answer pairs. \\
         
        \\
        
        \sethlcolor{mygreen}\hl{\textbf{\texttt{Text-only QA:}}} \\
        \quad Based on the above text, please design three question-answer pairs with different levels of difficulty: Easy, Medium, Hard.\\ 
        \quad The questions should be close-ended and should be answered based on the provided text. \\
        \quad The answer form should be as diverse as possible, including [Yes/No, Short Answer, Long Answer, Abstractive Answer]. \\
        \quad You should provide the reference in the text and the answer form if possible. \\
        \quad The output should be formalized as: '''Q: | A: | Reference: | Difficulty Level: | Answer Form:''' \\

        \\
        
        \sethlcolor{mygreen}\hl{\textbf{\texttt{Multimodal QA (w/table+text):}}} \\
        \quad Based on the above table and text, please design three question-answer pairs with different levels of difficulty: Easy, Medium, Hard.\\ 
        \quad The text provided is text related to the table, which can provide more reference for question generation, but the focus is still on the table itself. \\
        \quad These questions require locating the specific information, simple or complex calculations, comparisons, finding the maximum and minimum, reading across rows and columns, etc. \\
        \quad Note that these questions also need to be realistic. You should provide the reason if possible. \\
        \quad The output should be formalized as: '''Q: | A: | Reference: | Difficulty Level: | Answer Form:''' \\
        
        \\

         \sethlcolor{mygreen}\hl{\textbf{\texttt{Multimodal QA (w/figure+text):}}} \\
        \quad Based on the above figure and text, please design three question-answer pairs with different levels of difficulty: Easy, Medium, Hard.\\ 
        \quad The text provided is text related to the figure, which can provide more reference for question generation, but the focus is still on the figure itself. \\
        \quad These questions require a deep reading of the meaning of the image. \\
        \quad Note that these questions also need to be realistic. You should provide the reason if possible. \\
        \quad The output should be formalized as: '''Q: | A: | Reason: | Difficulty Level: | ''' \\

        \\

        \sethlcolor{mygreen}\hl{\textbf{\texttt{Multimodal QA (w/table):}}} \\
        \quad Based on the above image, please design three question-answer pairs with different levels of difficulty: Easy, Medium, Hard.\\ 
        \quad These questions require locating the specific information, simple or complex calculations, comparisons, finding the maximum and minimum, reading across rows and columns, etc. \\
        \quad Note that these questions also need to be realistic. You should provide the reason if possible. \\
        \quad The output should be formalized as: '''Q: | A: | Reason: | Difficulty Level: | ''' \\
        
        \\

        \sethlcolor{mygreen}\hl{\textbf{\texttt{Multimodal QA (w/figure):}}} \\
        \quad Based on the above image, please design three question-answer pairs with different levels of difficulty: Easy, Medium, Hard.\\ 
        \quad These questions require a deep reading of the meaning of the image.
        \quad  Note that these questions also need to be realistic. You should provide the reason if possible. \\
        \quad The output should be formalized as: '''Q: | A: | Reason: | Difficulty Level: | ''' \\
\bottomrule
\end{tabular}
\end{table*}
\endgroup

\section{Performance Comparison}
\begin{figure}[t]
\centering
\includegraphics[width=0.98\textwidth]{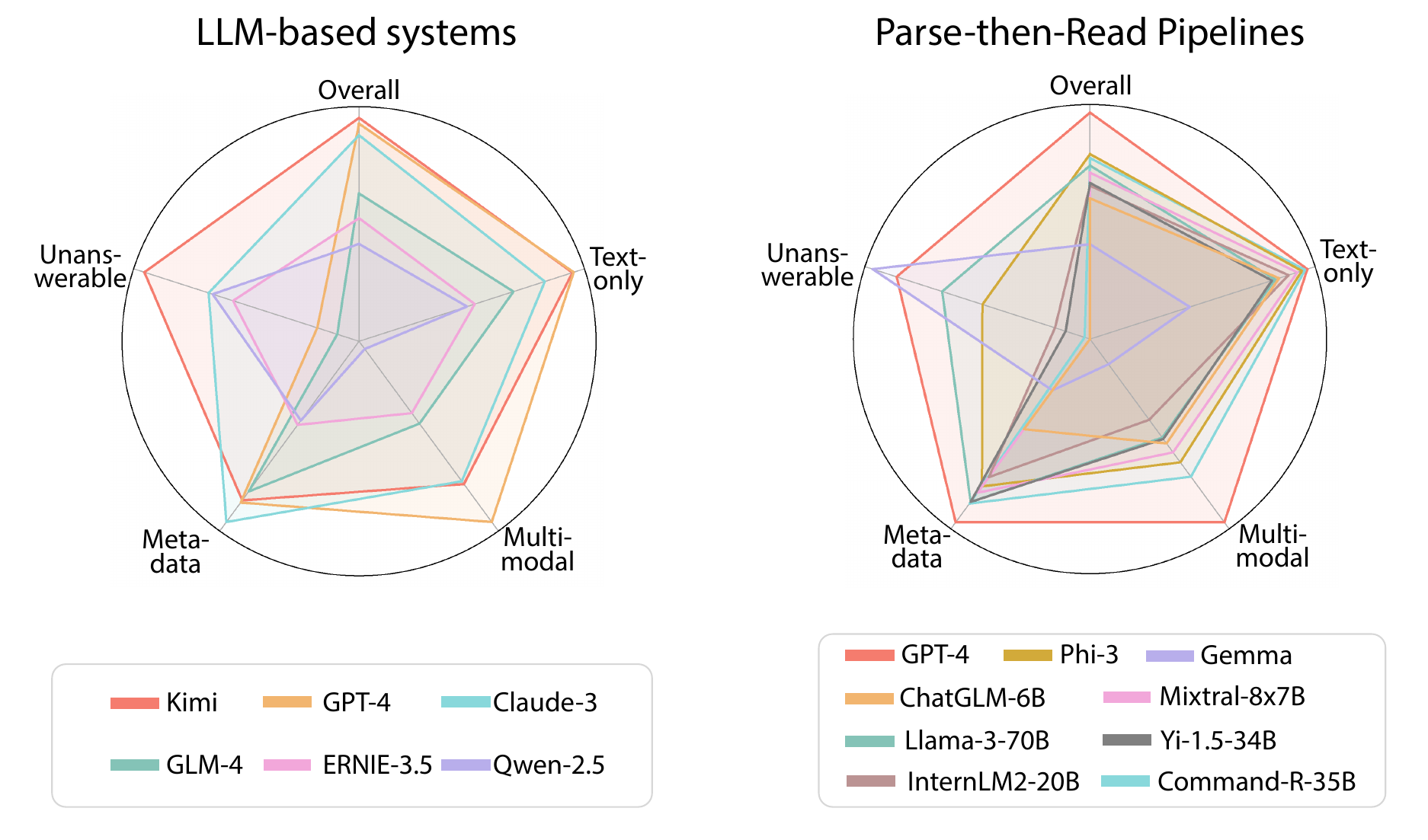}
\caption{Performance (Relative) of two major methods on \textsc{DocBench} against the best.}
\label{fig:radar}
\end{figure}

Figure \ref{fig:radar} demonstrates the relative performance of LLM-based systems and parse-then-read pipelines against the best on \textsc{DocBench}. 
For LLM-based systems, KimiChat consistently scores high across various metrics, demonstrating balanced performance. Notably, GPT-4 performs poorly in the unanswerable category, indicating potential overfitting in optimized GPT-4 file systems, which leads to decreased fidelity to given documents. Additionally, Claude-3 excels in the meta-data category, highlighting its superior ability to comprehend high-level metadata information.
For parse-then-read pipelines, we select models with the highest overall accuracy for comparison. Unlike LLM-based systems, GPT-4 demonstrates consistently high and balanced performance across all aspects within this pipeline. Notably, significant discrepancies arise in handling multi-modal and unanswerable questions, where GPT-4 and Gemma exhibit clear distinctions from the remaining methods.

\section{Analysis of Input Sources}
Table \ref{tab:analysis_input} presents the impact of different input sources on model performance. We provide questions to GPT-4 and GPT-4o, both with and without attached files. Remarkably, even without files, the models correctly answer a portion of the questions (19.1\% for GPT-4 and 21.7\% for GPT-4o). Our analysis reveals that the correctly answered questions are predominantly textual and are largely associated with government, law, and news domains. This trend suggests that the models' underlying training data is heavily skewed towards these categories, enabling them to answer some questions accurately without additional files. Moreover, as GPT-4o is an optimized version of GPT-4, it likely benefits from a broader and more extensive training data.

\begin{table*}[htbp]
\centering
\caption{Analyzing the Influence of Input Sources: We deliver questions with attached files and without files to GPT-4 and GPT-4o for evaluation, respectively.}
\footnotesize
\renewcommand\arraystretch{1.2}
\label{tab:analysis_input}
\resizebox{\linewidth}{!}{%
\begin{tabular}{lcccccccccc}
\toprule
\multirow{2}{*}{\textbf{Methods}}  & \multicolumn{5}{c}{\textbf{Domain}} & \multicolumn{4}{c}{\textbf{Type}} & \multirow{2}{*}{\textbf{Overall $\mathrm{Acc.}$}} \\ 
\cmidrule(lr){2-6} \cmidrule(lr){7-10} 
&\textbf{Aca.} & \textbf{Fin.} & \textbf{Gov.} & \textbf{Laws} & \textbf{News} &\textbf{Text.} & \textbf{Multi.} & \textbf{Meta.} & \textbf{Una.} & \\
\midrule
GPT-4    \\
\quad w/ file &65.7 &65.3 &75.7 &69.6 &79.6 &87.9 &74.7 &50.8 &37.1 &69.8  \\
\quad w/o file &10.9 &10.8 &23.0 &29.3 &32.6 &40.8 &8.1 &1.6 &10.5 &19.1  \\
\midrule
GPT-4o   \\
\quad w/ file
&56.4 &56.3 &73.0 &65.5 &75.0 &85.0 &62.7 &50.4 &17.7 &63.1  \\ 
\quad w/o file &11.2 &13.5 &29.1 &31.9 &36.0 &46.6 &10.7 &2.3 &6.5 &21.7  \\
\bottomrule
\end{tabular}
}
\end{table*}

\end{document}